\documentclass[unnumsec,webpdf,contemporary,large]{oup-authoring-template}%
\makeatletter
\@numbibtrue   
\setcitestyle{numbers}   
\PassOptionsToPackage{numbers}{natbib}
\makeatother
\usepackage{booktabs}

\graphicspath{{Fig/}}

\theoremstyle{thmstyleone}%
\theoremstyle{thmstyletwo}%
\theoremstyle{thmstylethree}%

\begin{document}

\journaltitle{Article}
\copyrightyear{2026}
\pubyear{2026}
\access{Advance Access Publication Date: Day Month Year}
\appnotes{Paper}

\firstpage{1}


\title[Short Article Title]{TwinPurify: Purifying gene expression data to reveal tumor-intrinsic transcriptional programs via self-supervised learning}

\author[1]{Zhiwei Zheng}
\author[1,$\ast$]{Kevin Bryson\ORCID{0000-0002-1163-6368}}

\authormark{Zheng et al.}

\address[1]{\orgdiv{School of Computing Science}, 
             \orgname{University of Glasgow}, 
             \postcode{G12 8RZ}, 
             \state{Glasgow}, 
             \country{United Kingdom}}

\corresp[$\ast$]{Corresponding author: 
\href{mailto:Kevin.Bryson@glasgow.ac.uk}{Kevin.Bryson@glasgow.ac.uk}}




\abstract{Advances in single-cell and spatial transcriptomic technologies have transformed tumor ecosystem profiling at cellular resolution. However, large scale studies on patient cohorts continue to rely on bulk transcriptomic data, where variation in tumor purity obscures tumor-intrinsic transcriptional signals and constrains downstream discovery. Many deconvolution methods report strong performance on synthetic bulk mixtures but fail to generalize to real patient cohorts because of unmodeled biological and technical variation. Here, we introduce TwinPurify, a representation learning framework that adapts the Barlow Twins self-supervised objective, representing a fundamental departure from the deconvolution paradigm. Rather than resolving the bulk mixture into discrete cell-type fractions, TwinPurify instead learns continuous, high-dimensional tumor embeddings by leveraging adjacent-normal profiles within the same cohort as “background” guidance, enabling the disentanglement of tumor-specific signals without relying on any external reference. Benchmarked against multiple large cancer cohorts across RNA-seq and microarray platforms, TwinPurify outperforms conventional representation learning baselines like auto-encoders in recovering tumor-intrinsic and immune signals. The purified embeddings improve molecular subtype and grade classification, enhance survival model concordance, and uncover biologically meaningful pathway activities compared to raw bulk profiles. By providing a transferable framework for decontaminating bulk transcriptomics, TwinPurify extends the utility of existing clinical datasets for molecular discovery.}


\maketitle

\begin{figure*}[!b]
\centering
\begin{minipage}{0.97\textwidth}
\footnotesize
\raggedright

\textbf{Zhiwei Zheng} is a PhD candidate at the University of Glasgow, researching deep learning for transcriptomics, GSEA optimization, and visualization methods for high-dimensional biological data.

\smallskip

\textbf{Kevin Bryson} is a Senior Lecturer at the University of Glasgow, focusing on agentic AI for integrative cancer bioinformatics and knowledge-guided analysis of high-throughput omics data.
\end{minipage}
\end{figure*}


\section{Introduction}
Bulk transcriptomic profiling remains the primary approach for measuring gene expression in large patient cohorts, despite the rapid rise of single-cell and spatial transcriptomics technologies that offer higher resolution and spatial information~\cite{scrnaseq, spatial}. While these new approaches provide unprecedented opportunities to study cellular heterogeneity and tissue architecture, they are often costly, technically challenging, and limited in scale. In contrast, bulk RNA-seq is broadly accessible, routinely applied in research and clinical settings, and serving as the foundation for large scale transcriptomic analyses across diverse biological contexts. 

However, the cellular heterogeneity of bulk tissues introduces substantial variation in tumor purity, which refers to the proportion of cancer cells within a sample.  Purity variation arises from immune and stromal infiltration, as well as the presence of normal tissue in the biopsy,  and can profoundly distort transcriptomic analyses. Unadjusted purity skews pathway enrichment results, biases molecular subtype assignments, reduces sensitivity for detecting low frequency variants, and undermines the reproducibility of prognostic models~\cite{Estimate, ISOpure, Purity_DNA_methylation, tumor_purity_microenvir}. Purity fluctuations can obscure prognostic signatures, weaken subtype stratification, and alter associations between biomarkers and treatment response~\cite{BCPS_2024, luminal_purity}. From a computational perspective, this confounding effect acts as a non-linear distortion on the feature space, obscuring subtle biological programs and altering associations between biomarkers and treatment response.  These challenges highlight the need for computational strategies that can recover cell-type-specific signals directly from bulk data to support reliable downstream molecular and clinical analyses.  

Several classes of computational methods have been developed to mitigate the confounding effects of tumor purity. No-reference approaches (e.g., PCA/NMF) attempt to infer latent components directly from bulk data, but operate without biological anchors and therefore recover factors that rarely correspond to true cell-type programs. Reference-based approaches can be further divided into two categories. The first includes fixed-signature methods, such as ESTIMATE~\cite{Estimate} and CIBERSORT~\cite{Cibersort}, which rely on predefined stromal or immune gene signatures and thus cannot adapt to cohort-specific biology. The second includes single-cell--assisted methods, such as MuSiC~\cite{Music} and TAPE~\cite{TAPE}, which train predictive models on pseudo-bulk mixtures generated from scRNA-seq reference atlases to infer cell-type--specific expression. DNA-based tools such as ABSOLUTE~\cite{Absolute} estimate tumor purity and ploidy from somatic copy-number alterations and allelic fractions, but require genomic rather than transcriptomic input and therefore fall outside expression-based deconvolution.

Despite methodological differences, all reference-based strategies operate on unpaired datasets and implicitly assume that fixed signatures or single-cell--derived pseudo-bulk models are transferable across heterogeneous cohorts. This assumption often fails due to substantial distributional gaps between reference and target data, including differences in cell-type composition~\cite{AvilaCobos2020}, sequencing depth and library strategy~\cite{Lu2025}, ambient RNA contamination~\cite{Young2020}, and technical or batch effects~\cite{Yu2024}. These mismatches lead to poor generalization and inaccurate recovery of tumor-intrinsic expression programs in real-world bulk transcriptomic samples, underscoring the need for scalable frameworks that can learn transferable tumor-specific representations directly from bulk data.

Here, we introduce TwinPurify (TP) as a novel external-reference-free strategy, a self-supervised learning framework designed to extract purified embeddings directly from bulk gene expression data. Representing a paradigm shift from deconvolution, TwinPurify leverages adjacent-normal samples as structured perturbations, enabling the model to disentangle cell-type-specific components without relying on predefined references. The framework is implemented as a Python package, enabling direct application to bulk transcriptomic datasets. We evaluated TwinPurify on three independent, large scale cancer cohorts comprising both RNA-seq and microarray data, generating controlled dilution series to benchmark its ability to recover target signals under varying purity levels. Across tasks including molecular subtyping, histological grade prediction, survival modeling, and functional pathway analysis, TwinPurify consistently outperformed conventional autoencoder-based models~\cite{AE, VAE} and demonstrated superior robustness to purity variation. Importantly, the learned embeddings captured distinct biological programs such as immune response and cell cycle regulation, highlighting the interpretability of the representations. Together, these advances establish TwinPurify as a scalable and biologically grounded framework for purifying bulk transcriptomes, enhancing the fidelity of molecular discovery and clinical modeling.


\section{Materials and methods}\label{sec2}

\subsection{Datasets and preprocessing}

We utilized three large-scale breast cancer cohorts: SCAN-B~\cite{scanb_dataset} and TCGA-BRCA~\cite{ucsc_xena_tcga_gtex} (RNA-seq), and METABRIC~\cite{ega_breast_cancer} (microarray). To ensure cross-platform consistency, gene identifiers were mapped to Ensembl IDs, and a unified set of 16,736 protein-coding genes shared across all platforms was retained. Expression values were standardized via a $\log_2(x + 1)$ transformation. Importantly, RNA-seq and microarray data were processed independently to prevent technical artifacts.

For model training and evaluation, cohorts were split into independent training and testing sets. SCAN-B splits followed the primary study definitions~\cite{scanb_dataset}; METABRIC was divided by discovery (train) and validation (test) cohorts; and TCGA was randomly split (80:20). Samples lacking PAM50 or clinical annotations were excluded. "Normal-like" samples were retained exclusively in test sets to assess performance on low-purity tumors without confounding training. Adjacent-normal tissue samples (denoted as "Pure Normal") were included in all cohorts to guide the self-supervised learning process. Detailed descriptions of sample filtering, normalization, and cohort statistics are provided in Supplementary Information.

\subsection{Self‑supervised framework and model Architecture}\label{subsubsec1}

We developed a self-supervised framework TwinPurify (TP) based on the Barlow Twins objective (Fig.~\ref{fig:BT_diagram}) to learn disentangled, cancer-specific representations from bulk transcrip\-tomics. The core idea is to guide the model to identify and remove normal-tissue signals embedded in tumor expression profiles. To this end, we introduced synthetic perturbations during training by simulating a contamination process. At each training epoch, the expression vector of a tumor sample is further mixed with additional normal profiles to simulate realistic admixtures. Specifically, each distortion is constructed as a convex combination of five randomly selected adjacent-normal samples (drawn without replacement from 46 SCAN-B normals in training set), with coefficients sampled uniformly to sum to one. Combining multiple samples mitigates the risk of contamination in individual adjacent-normals, while selecting five balances sample diversity without excessive repetition. This augmentation strategy replaces conventional additive Gaussian noise with structured, physiologically grounded perturbations, compelling the network to detect and suppress normal-cell expression signals that are not intrinsic to the tumor.

To generate robust contrastive pairs, each training sample is first constructed from a controlled mixture of tumor and normal signals. For a given raw tumor expression vector $\mathbf{x}^{(\mathrm{tumor})}$, to mitigate potential contamination in adjacent–normal profiles, we create a synthetic normal reference by randomly sampling five distinct normal profiles $\{\mathbf{n}_1,\ldots,\mathbf{n}_5\}$ and mixing them with randomly drawn proportions $\{\alpha_k\}_{k=1}^5$ such that $\sum_{k}\alpha_k = 1$:

\begin{align}
\mathbf{x}^{(\mathrm{normal})}
= \sum_{k=1}^{5} \alpha_k \, \mathbf{n}_k .
\end{align}

Each training input is constructed by mixing the tumor expression with a synthetic normal background at a fixed tumor-to-normal ratio. The mixing coefficient is selected via Optuna from a predefined search range:
\begin{align}
\mathbf{x}
= \alpha \, \mathbf{x}^{(\mathrm{tumor})}
\;+\;
(1 - \alpha) \, \mathbf{x}^{(\mathrm{normal})}.
\end{align}
Here, $\alpha$ denotes the tumor mixing coefficient chosen through hyperparameter optimization; in our experiments, the optimal value was $\alpha = 0.27$ (tumor) and $1-\alpha = 0.73$ (normal).

We created two independent views for representation learning, they have different normal expression sampling in the above optimal ratio.

\begin{align}
\tilde{\mathbf{x}}_1 = \mathcal{T}_1(\mathbf{x}), 
\qquad
\tilde{\mathbf{x}}_2 = \mathcal{T}_2(\mathbf{x}).
\end{align}

Both views are passed through a shared encoder $f_{\theta}$ and a projector head $g_{\phi}$, producing two projection outputs:

\begin{align}
\mathbf{z}_1 = g_{\phi}(f_{\theta}(\tilde{\mathbf{x}}_1)),
\qquad
\mathbf{z}_2 = g_{\phi}(f_{\theta}(\tilde{\mathbf{x}}_2)).
\end{align}

The projection vectors $\mathbf{z}_1$ and $\mathbf{z}_2$ are first normalized along the batch dimension to have zero mean and unit variance. Given a batch of paired projection outputs, we compute the cross-correlation matrix
$\mathbf{C} \in \mathbb{R}^{d \times d}$:

\begin{align}
C_{ij}
= 
\frac{
\sum_{b=1}^{B}
z_{1,i}^{(b)} \, z_{2,j}^{(b)}
}{
\sqrt{
\sum_{b=1}^{B} \big(z_{1,i}^{(b)}\big)^{2}
}
\sqrt{
\sum_{b=1}^{B} \big(z_{2,j}^{(b)}\big)^{2}
}
}.
\end{align}

Here, $b$ indexes samples in the mini-batch ($b=1,\dots,B$), where $B$ is the batch size. The notation $z_{1,i}^{(b)}$ denotes the $i$-th dimension of the projection vector from view 1 for the $b$-th sample in the batch (similarly for $z_{2,j}^{(b)}$).

We then compute the cross-correlation matrix, the tumor-preserving (TP) objective encourages invariance across distortions (diagonal terms $\to 1$) while enforcing decorrelation across latent dimensions (off-diagonals $\to 0$). Formally, given the cross-correlation matrix $\mathbf{C} \in \mathbb{R}^{d \times d}$ between two distorted embeddings, the TP loss is defined as:

\begin{align}
\mathcal{L}_{\mathrm{TP}}
=
\sum_{i} \left(1 - C_{ii}\right)^{2}
\;+\;
\lambda \sum_{i} \sum_{j \ne i} C_{ij}^{2}.
\end{align}

In our implementation, $\lambda$ was tuned within the range of 10-100, and the final selected value was $54.9$, ensuring an appropriate balance between the diagonal and off-diagonal constraints. By driving the embedding space toward an identity cross-correlation structure, the model retains tumor-specific biological signals while suppressing redundant or confounding normal components, resulting in a representation that is robust to noise and potential contamination.

To evaluate this approach, we compared TP with two canonical baselines: a standard autoencoder (AE) and a variational autoencoder (VAE), as well as PCA (Principal Component Analysis) as a non-deep-learning baseline. All deep learning models were trained on the same input structure: each tumor profile and its adjacent-normal were used to form a composite input vector. However, only TP incorporated additional distortions, which is constructed by mixing in randomly sampled adjacent-normal profiles at every training step. These distortions were designed to simulate varying degrees of normal-cell contamination and encourage the model to learn tumor-specific representations by actively removing normal signals. By contrast, AE and VAE operated without explicit distortions and relied purely on reconstruction-based objectives. For PCA, input features were adjacent in dimensionality to the deep learning models and fitted on the training set, then transformed on the test set. Together, these baselines allowed us to assess the effectiveness of the TP objective in disentangling oncogenic signals under biologically realistic perturbations.

\subsection{Dilution evaluation experiment design}\label{subsec2}

To demonstrate that TP faithfully recovers tumor signals from bulk expression data, and that higher purity enhances downstream analyses, we generated synthetic mixtures of tumor and adjacent-normal transcriptomics across a spectrum of dilutions (from 0\% to 100\% tumor content in 10\% increments). By incrementally increasing the dilution rate, we could directly observe the robustness of downstream tasks (e.g., classification) under progressively reduced cancer signal. Accordingly, for the three breast cancer cohorts, we evaluated two canonical endpoints at each dilution: (i) Concordance of PAM50 intrinsic subtypes with reference labels provided by the original studies; and (ii) Histological grade accuracy. Note that the TCGA dataset does not have grade annotations.

Hyperparameter optimization for each encoder (AE, TP, and VAE) was carried out using Optuna~\cite{optuna_akiba2019}. The best‐performing encoders were then applied to the original bulk expression profiles to derive low-dimensional embeddings for subsequent analyses. After training the encoders on the original bulk gene expression data, the learned weights were fixed. For each downstream classification task, we first encoded the undiluted training samples through the pretrained encoder to obtain embeddings. Using these embeddings, we trained a multinomial logistic regression classifier under stratified 5-fold cross-validation. Within each training fold, a portion of the data was held out as an internal validation set, resulting in five fold-specific models. 

For PCA, we first fit the linear transformation on the training set to match the dimensionality of the encoder embeddings. The training samples were projected through PCA to obtain embeddings, and the same 5-fold cross-validation logistic regression procedure was applied.

For evaluation on synthetic dilutions, all dilution-specific test samples were projected through the corresponding pretrained encoder (for AE, TP, VAE) or the fitted PCA transformation. Each of the five fold-specific logistic regression models was then used to predict the class labels for the test embeddings. Final predictions were obtained by majority vote across the five models, providing an ensemble-based estimate of classification performance for each dilution level.

Throughout all figures presenting model predictions, error bars reflect variability across the 5 cross-validation folds, representing uncertainty due to differences in training data splits rather than sample-level variation.

\subsection{Biological validation: GSEA and survival prediction pipelines}
To interrogate the biological meaning of the embedding space and test the hypothesis that TP produces orthogonal features, we implemented the following pipeline. For each embedding dimension \(d_i\), we calculated the Pearson correlation coefficient between \(d_i\) and the expression level of every gene in the original bulk dataset, thus obtaining a ranked list of genes ordered by their correlation with \(d_i\). Repeating this procedure for all \(n\) dimensions yielded \(n\) preranked gene lists. Each preranked list served as input to the GSEA Java Desktop in prerank mode. We tested gene-set enrichment against the GO Biological Process (GO--BP) collection, using a minimum gene-set size of 15, a maximum of 500, and 1,000 gene-set permutations. This approach generated dimension-specific GSEA profiles, enabling us to map each orthogonal embedding axis to distinct biological pathways and processes.

Before generating the dimension-pathway heatmaps, we first performed a comprehensive pathway-level analysis across multiple models. Specifically, we evaluated four embedding-based methods, AE, TP, VAE, and BT-with-Gaussian-noise (a Barlow Twins variant trained with Gaussian noise instead of adjacent-normals), to assess the number and diversity of enriched pathways. For each model, we collected all pathways with (FWER $<$  0.05 / number of dimensions) from GSEA results, applying a Bonferroni correction to account for multiple testing, and counted unique pathways across dimensions, so that duplicate pathways appearing in multiple dimensions were counted only once per model. For PCA, the preranked gene list was constructed according to gene loadings ordered by each principal component’s contribution, followed by GSEA to identify significantly enriched pathways (FWER~$<$  0.05). For the differential expression (DE) approach, we performed two-sample t-tests between tumor and adjacent-normal samples, using the t-statistics as the preranked metric for GSEA. Additionally, two external GSEA implementations, GOAT and BlitzGSEA were evaluated via their web-based interfaces, both using the DE-derived gene ranking. Across all analyses, we used consistent parameters: gene-set sizes between 15 and 500, and 1,000 phenotype permutations. The resulting enrichment statistics were summarized into two bar plots representing the number of significant pathways in the GO--BP and Immunologic Signature (C7) collections, respectively.

To quantify the independence of learned representations, we defined a dimension-wise uniqueness score $u_k$ based on the overlap of top-ranked genes between latent dimensions. Statistical significance was assessed via a permutation test (see Supplementary Information for full mathematical derivation and definitions).

Subsequently, for pathway-level orthogonality, we retained only GO--BP terms with available normalized enrichment scores (NES) and with family-wise error rate (FWER~$<$  0.01), ensuring that only statistically significant and properly computed enrichment results were considered. Pathways without NES were excluded because the enrichment score could not be reliably calculated, rendering downstream similarity analyses biologically meaningless. To reduce redundancy among enriched GO--BP terms, we computed pairwise semantic similarities using the Wang measure~\cite{GOSEMSIM_measure} implemented in GOSemSim~\cite{GOSEMSIM} and performed hierarchical clustering (average linkage) on the distance metric \(1 - \text{similarity}\). Clusters were cut at a semantic-similarity threshold of 0.3, yielding a concise set of nonredundant GO themes for each latent axis. We then selected the top five positively and top five negatively enriched signatures per axis and visualized their NES values in a heatmap, facilitating comparison of the strongest GO signals across models and dimensions. The same correlation-derived GSEA results were subsequently used to assess enrichment of immunologic signature gene sets, applying the same filtering criteria of NES availability and FWER~$<$  0.01.

Finally, to examine whether the gene sets derived from each model capture distinct and clinically relevant biological signals, we conducted a survival analysis based on model-specific gene lists. For AE, VAE, TP, and BT-with-noise, we selected the top 20 and bottom 20 genes from each of the four embedding dimensions, resulting in up to 160 unique genes per model (fewer in cases of overlap). For PCA and DE analyses, the top 80 and bottom 80 genes were selected directly according to their loadings or t-statistics, respectively. As BlitzGSEA and GOAT relied on the same DE-based rankings, they were excluded from this step. For each model, the selected genes were intersected with the original bulk expression matrix, and the resulting subset was used to fit a Cox proportional hazards model to predict survival risk scores. Samples were stratified into high- and low-risk groups according to the median risk, followed by Kaplan-Meier (KM) survival analysis and log-rank testing. The concordance index (C-index) was computed to quantify survival prediction performance across models.

\section{Results}

\subsection{Overview of TwinPurify}

\begin{figure*}[htbp]
  \centering
  \includegraphics[width=\textwidth]{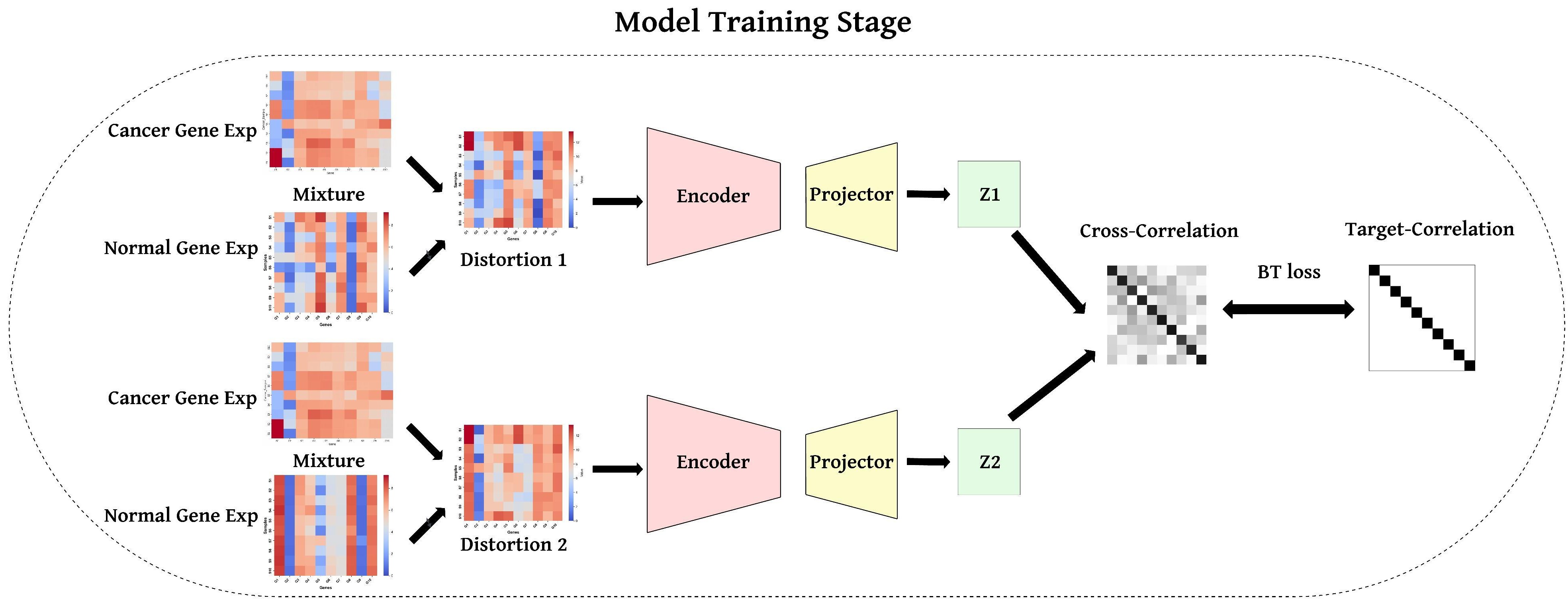}
  \caption{%
    \textbf{TwinPurify embedding workflow for generating purified cancer gene expression embeddings.} 
    Cancer bulk‑RNA profiles are synthetically “contaminated” with varying fractions of adjacent-normal tissue expression to produce two perturbed views. Both views are passed through a shared encoder aligned via the Barlow‑Twins loss. }
  \label{fig:BT_diagram}
\end{figure*}

The TwinPurify (TP) framework is a external-reference-free representation learning pipeline designed to disentangle tumor-specific transcriptional signals from the confounding adjacent normal tissue background (Figure \ref{fig:BT_diagram}).  TP adopts an encoder–projector architecture inspired by the Barlow Twins~\cite{BT} objective, adapted to operate on gene expression profiles rather than images. To learn more tumor-specific embeddings, we introduced directed noise using adjacent-normal tissue profiles, enabling the model to separate tumor signals from background more effectively. All models were pre-trained on SCAN-B breast cancer cohort and fine-tuned on METABRIC and TCGA cohorts. TP embeddings were applied to multiple downstream tasks, including PAM50 subtyping and histological grade prediction, using multinomial logistic regression for classification. For further biological validation, individual embedding dimensions were correlated with gene expression to generate preranked gene lists. For each list, the top and bottom genes were selected as signatures for survival prediction and further analyzed using GSEA to uncover enriched biological pathways. TP was benchmarked against two reconstruction-based autoencoder baselines (AE and VAE) and a conventional dimensionality reduction method (PCA~\cite{PCA}), providing a robust representation for downstream predictive and biological analyses. 

\subsection{Robustness of PAM50 intrinsic subtype prediction under dilution}
\begin{figure*}[htbp]
  \centering
  \includegraphics[width=0.8\linewidth]{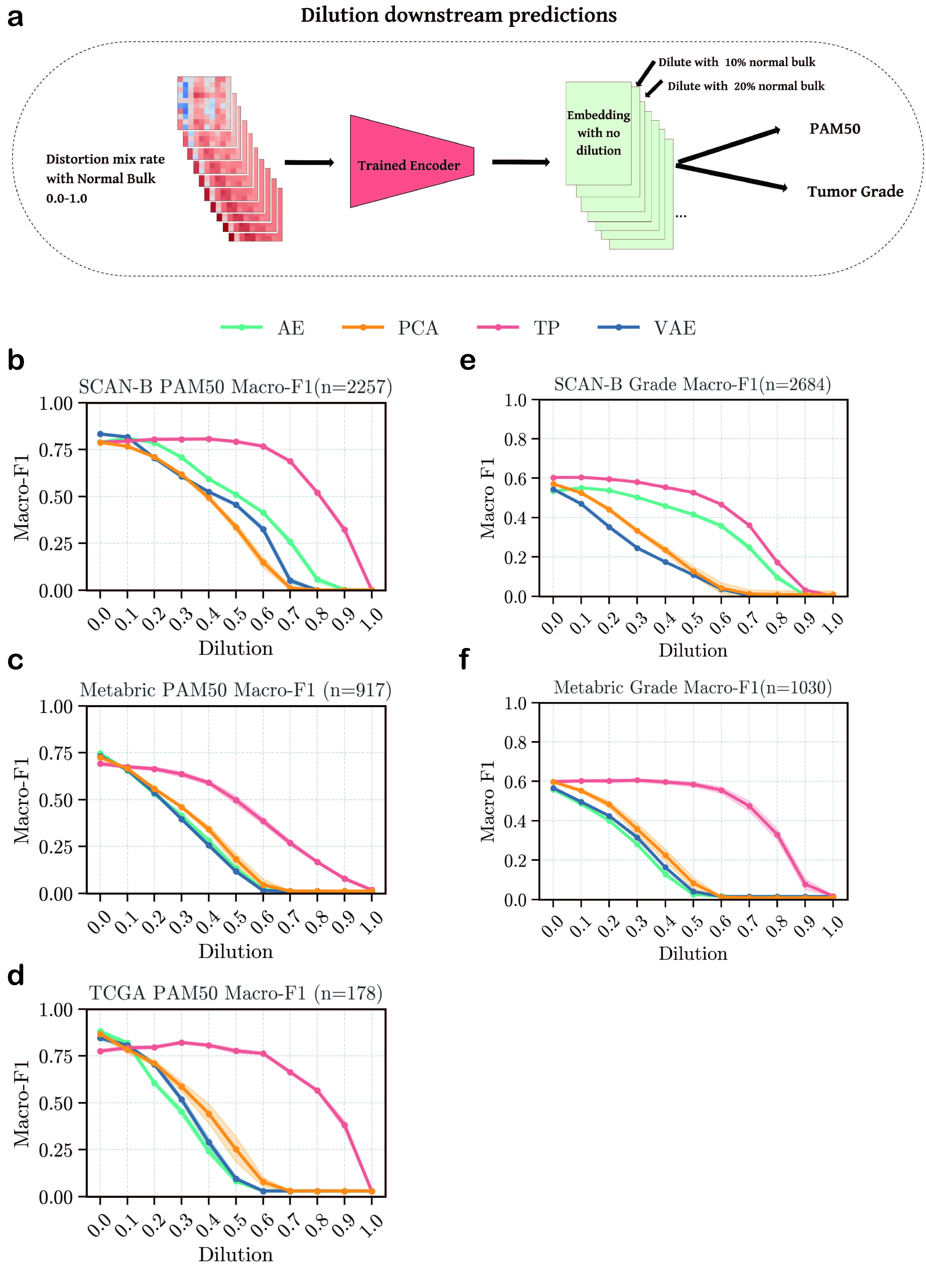}
  \caption{%
    \textbf{Downstream predictions under increasing adjacent-normal distortion.}
    \textbf{a}. The trained encoder maps held-out cancer samples with real normal admixture into the learned embedding space. These embeddings serve as input features for tumor-subtype classification and cancer grade prediction.
    \textbf{b-d}. Macro-F1 scores of PAM50 subtype classification for the SCAN-B, METABRIC, and TCGA cohorts, respectively, as the dilution rate increases, across four models: autoencoder (AE), variational autoencoder (VAE), TwinPurify (TP), and principal component analysis (PCA).
    \textbf{e-f}. Macro-F1 scores of cancer grade classification for the SCAN-B and METABRIC cohorts as the dilution rate increases, evaluated using the same four models.
  }
  \label{fig:Dilution_line_charts}
\end{figure*}

Here, we evaluated whether TP embeddings preserve intrinsic subtype boundaries under increasing normal-tissue contamination. PAM50 classification~\cite{PAM50} is a well-established, clinically validated assay for intrinsic breast cancer subtyping and serves as a key benchmark of molecular fidelity. The five canonical PAM50 subtypes (Basal-like, HER2-enriched, Luminal~A, Luminal~B, and Normal-like) reflect distinct transcriptional programs and clinical behaviors. Due to the ambiguous boundary between Normal-like and other tumor subtypes~\cite{Normal_like1, Normal_like2}, we introduced pure adjacent-normal (PN) samples as an external reference to ensure clearer separation and more reliable evaluation. To assess the ability of TwinPurify (TP) to disentangle tumor-specific signals from normal background, we evaluated PAM50 subtype classification using embeddings derived from synthetic mixtures. These mixtures were created by blending tumor expression profiles with adjacent-normal samples at increasing ratios and then processed through the trained encoder (Fig.~\ref{fig:Dilution_line_charts}a) to obtain latent representations for classification (dilution~\(=0\) corresponds to pure tumor; dilution~\(=1\) corresponds to pure adjacent-normal). In the SCAN-B cohort (Fig.~\ref{fig:Dilution_line_charts}b), all models performed comparably on undiluted data but diverged sharply as normal contamination increased. When tumor signals were reduced by 60\%, TP still retained a Macro-F1 above 0.75, whereas AE, VAE, and PCA had already dropped below 0.5. At 90\% dilution, TP continued to produce non-random predictions, while the other models completely collapsed toward uniform PN classification.

The same robustness pattern was observed across cohorts: models trained and evaluated on METABRIC recapitulated TP’s superior resistance to dilution (Fig. ~\ref{fig:Dilution_line_charts}c), and evaluation on the smaller TCGA cohort (Fig. ~\ref{fig:Dilution_line_charts}d) showed qualitatively similar results, indicating that TP’s advantage generalizes across different dataset types and sample sizes. Together, these observations indicate that TP more effectively preserves tumor-relevant signal in the presence of increasing normal contamination compared with reconstruction-based autoencoders and PCA.

\begin{figure*}[htbp]
  \centering
  \includegraphics[width=\linewidth]{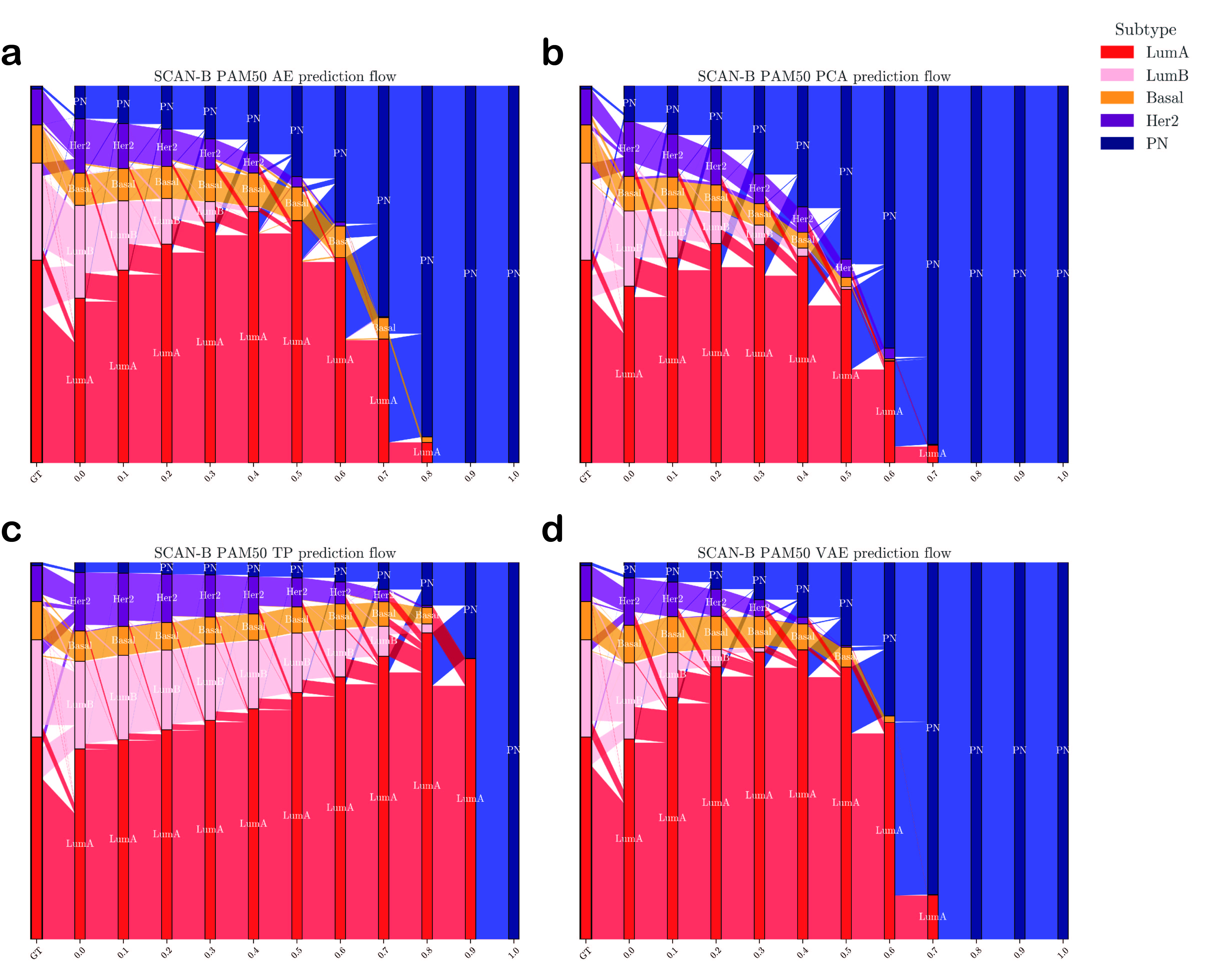}
  \caption{{%
    \textbf{PAM50 subtype transitions under increasing adjacent-normal distortion in the SCAN-B cohort.} 
    Plots depict how PAM50 subtype predictions for SCAN-B test samples (n=2257) change across different dilution levels. GT indicates the original ground truth labels, and subsequent bars show predictions under increasing distortion. Results are shown for four models: autoencoder (AE), variational autoencoder (VAE), TwinPurify (TP), and principal component analysis (PCA).
    }
  }
  \label{fig:PAM50 subtype transitions}
\end{figure*}

To further examine sample-level consistency beyond the averaged F1 trends, we visualized the prediction dynamics across dilution levels using per sample trajectory plots (Fig. ~\ref{fig:PAM50 subtype transitions}), which provides an intuitive visualization of how each sample’s predicted subtype evolves across dilution levels, complementing the aggregate trends observed in the line plots. Overall, TP exhibited a markedly higher sensitivity to tumor-normal contrast: even at extreme dilution (90\%), most samples remained classified as LumA, whereas the other methods had largely collapsed into the pure normal (PN) state. In other words, TP’s trajectories predominantly showed subtype retention or gradual transitions, while AE, VAE, and PCA displayed rapid and complete convergence toward PN. This sample-level stability further supports the robustness indicated by the aggregate metrics. TP preserves biologically meaningful subtype boundaries even under strong normal background interference, effectively disentangling tumor-specific signals from normal noise. At the subtype level, TP demonstrated distinct robustness patterns compared with the other models. For the Basal-like and HER2-enriched subtypes, all models exhibited a similar decreasing trend as dilution increased; however, AE, VAE, and PCA ceased predicting these aggressive subtypes as early as the 0.5-0.7 dilution range. In contrast, TP maintained accurate predictions up to 0.8-0.9 dilution, only gradually reassigning the remaining samples to the predominant LumA category. A similar pattern was observed for the LumB subtype, which disappeared in other models at around 0.3-0.4 dilution, while TP continued to identify a small proportion of LumB samples even at 0.8. Together, these results indicate that TP effectively amplifies tumor-normal distinctions, retaining biologically meaningful subtype boundaries under heavy contamination with adjacent-normal tissue. 

Our results highlight the nuanced nature of the LumA and LumB subtypes. As studies suggested~\cite{PAM50_discontinum, PAM50_CONTINUM1}, these subtypes may represent a phenotypic continuum rather than strictly discrete classes. In our experiments, all four models occasionally mis-classified ground truth LumA samples as LumB, illustrating the inherent similarity between these subtypes in expression space. With increasing dilution, we observed a progressive reassignment of LumB samples toward LumA, plausibly due to the dilution of proliferative signals such as Ki67~\cite{KI671}. Notably, TP was more resilient to this effect than AE or VAE, preserving subtle subtype distinctions even under high dilution. This behavior underscores the blurred boundary between LumA and LumB, supporting the view that they form a continuous spectrum rather than entirely separate entities. Importantly, the gradual LumB-to-LumA transition observed in all models further suggests that this is a biological feature rather than a model-specific artifact.

\subsection{Effect of dilution on clinical grade prediction robustness}

We evaluated model performance on histological tumor grade under increasing adjacent-normal admixture (Fig.~\ref{fig:Dilution_line_charts}e-f), using Grade 0 to represent pure adjacent-normal tissue. TP consistently outperformed AE, VAE, and PCA, achieving F$_1$ score of 0.603 (3--7\% higher than baselines) and maintaining F$_1 > 0.50$ even at 50\% dilution, whereas other models dropped below 0.40. While all methods showed reduced sensitivity for intermediate Grade 2 tumors, TP sustained higher relative accuracy for the more distinct Grades 1 and 3 (Supplementary Fig.~S1). Analysis of grade-transition trajectories (Supplementary Fig.~S2) further confirmed TP's robustness: while baseline models rapidly converged to Grade 0 (normal) predictions at 70--80\% dilution, TP decelerated this signal decay, retaining malignant signatures (predicting Grade $\ge$ 1) even at 90\% dilution. These results demonstrate that TP embeddings encode clinically meaningful variation that remains discernible under substantial contamination.

\subsection{Biological validation of learned representations via GSEA and survival prediction}

To validate biological interpretability, we systematically annotated latent axes using gene set enrichment analysis (GSEA)~\cite{GSEA1} (Fig.~\ref{fig:GSEA_bar_charts}a). Pearson correlations between each latent dimension and all input genes were computed to generate dimension-specific preranked lists. These rankings were then evaluated via (i) preranked GSEA to identify enriched biological pathways, and (ii) survival prediction using top- and bottom-ranked genes to assess the prognostic relevance of each latent axis.

\subsubsection{Pathway enrichment evaluation of latent dimensions} 

\begin{figure*}[htbp]
  \centering
  \includegraphics[width=\linewidth]{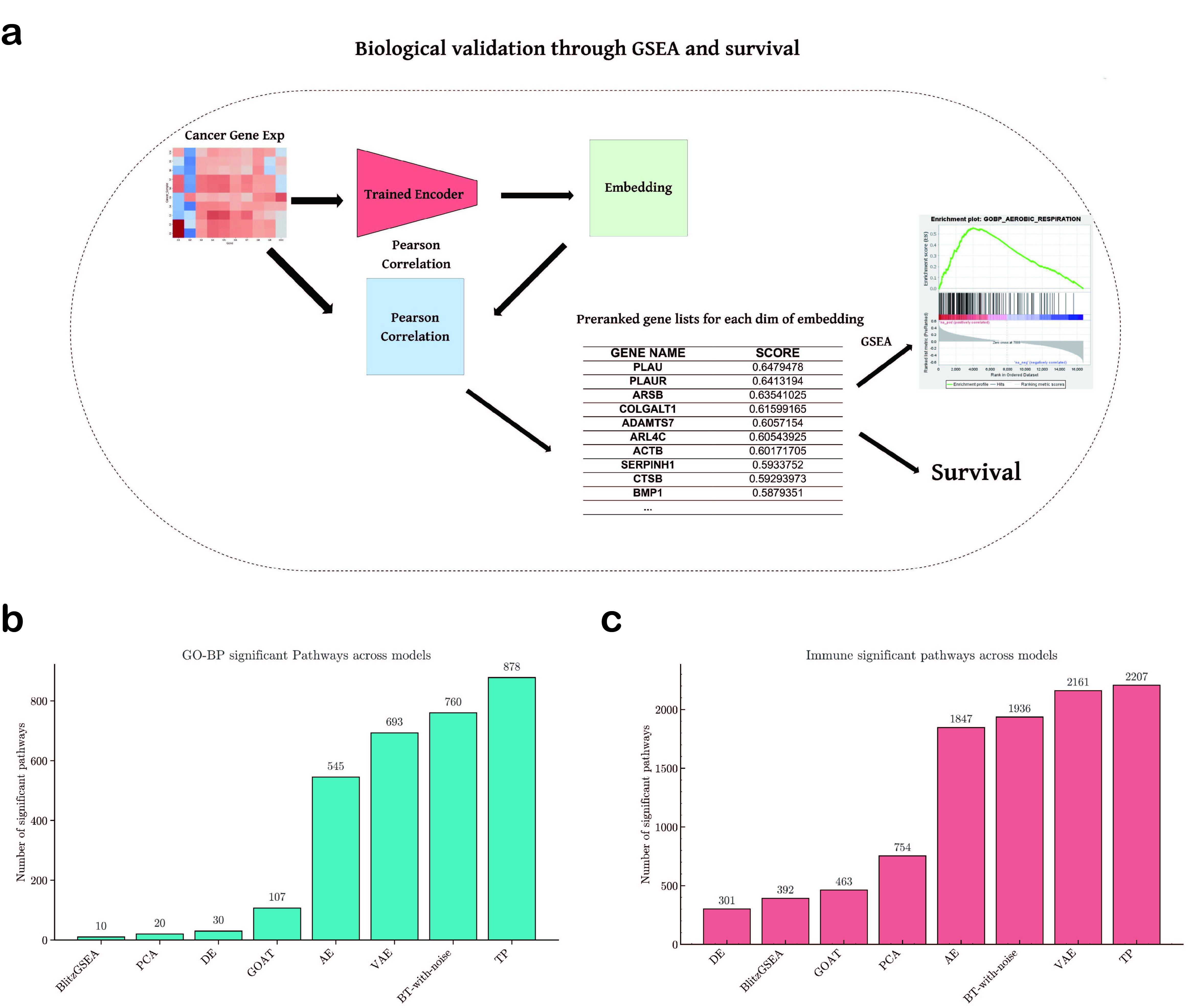}
  \caption{%
    \textbf{Biological validation of learned embeddings.}
    \textbf{a} Pearson correlation is computed between each learned dimension and the original (uncontaminated) cancer profiles. Preranked gene lists derived from correlated dimensions are analyzed by gene set enrichment analysis (GSEA) to recover known oncogenic pathways.
    \textbf{b} The number of significantly enriched pathways (FWER $<$   0.05/ number of dimensions for embedding-based and FWER $<$  0.05 for non-embedding based) within the GO Biological Process collection for each model: differnial gene analysis(DE), BlitzGSEA, BT-with-noise(Barlow-Twins with Gaussian noise), GOAT, autoencoder (AE), variational autoencoder (VAE), TwinPurify(TP), and principal component analysis (PCA).
    \textbf{c} The number of significantly enriched pathways with the same constraint like above, within the Immunologic Signature gene sets for the same eight models.
  }
  \label{fig:GSEA_bar_charts}
\end{figure*}

We applied GSEA to GO biological processes~\cite{GO1,GO2,GO3,GO4,GO5}, quantifying unique, Bonferroni-corrected pathways enriched across latent dimensions for TP and baselines (including PCA, BlitzGSEA \cite{BlitzGSEA}, GOAT \cite{GOAT}, and DE) (Fig.~\ref{fig:GSEA_bar_charts}b). TP identified substantially more significant pathways than all comparators. Notably, TP outperformed models trained with artificial Gaussian noise~\cite{SelfSupervised_NM}, demonstrating that biologically grounded perturbations (mixing adjacent-normal profiles), rather than generic noise, are essential for encouraging the encoder to learn axes aligned with true biological variation.

We further assessed immune signal preservation using the Immunologic Signature collection ~\cite{Immune_GSEA}. By quantifying significantly enriched pathways across frameworks (Fig.~\ref{fig:GSEA_bar_charts}c), we determined whether tumor purification compromised immune context. Instead, TP retained the level of immune-pathway enrichment observed in VAE and other baselines, and in several cases even exceeded it. This suggests that TP’s ability to reduce normal tissue signals not only enhances the purity of tumor-derived transcriptional programs but also preserves biologically relevant variation from the tumor microenvironment. This distinction arises because TwinPurify treats adjacent healthy tissue as the background noise to be removed, distinct from the tumor-infiltrating immune cells (TILs) which are integral to the tumor microenvironment. By contrasting tumor samples against “immune-quiet” adjacent-normal references, TP effectively suppresses the background healthy signal while preserving and computationally enhancing the specific signals of tumor-associated immune infiltration.

\subsubsection{Gene-level and pathway-level independence evaluation}

To validate the orthogonality of TP embeddings, we evaluated gene-level independence by computing pairwise overlaps between the top and bottom 1,000 genes correlated with each latent dimension. Statistical significance was established via 1,000 permutations. To control for random effects, we performed 1,000 permutations. TP consistently exhibited minimal overlap between gene lists from distinct dimensions, indicating strong independence (Table~\ref{tab:dim_independence}). In contrast, AE, VAE, and GN showed extensive overlap with non-significant permutation p-values, confirming that their latent dimensions do not capture independent transcriptional programs. These results demonstrate that TP embeddings effectively disentangle biologically distinct gene sets across latent axes, supporting both interpretability and downstream utility.

\begin{table*}[htbp]
\centering
\caption{Independence of correlation-based gene rankings across model dimensions. 
TP demonstrates significantly higher uniqueness (fraction of unique genes) compared to AE and VAE, indicating successful disentanglement of biological signals. Empirical p-values computed with 1,000 permutations.}
\label{tab:dim_independence}
\setlength{\tabcolsep}{8pt}
\begin{tabular}{lcccccccc}
\toprule
 & \multicolumn{2}{c}{\textbf{Dimension 0}} & \multicolumn{2}{c}{\textbf{Dimension 1}} & \multicolumn{2}{c}{\textbf{Dimension 2}} & \multicolumn{2}{c}{\textbf{Dimension 3}} \\
\cmidrule(lr){2-3} \cmidrule(lr){4-5} \cmidrule(lr){6-7} \cmidrule(lr){8-9}
\textbf{Model} & Unique & $P$-val & Unique & $P$-val & Unique & $P$-val & Unique & $P$-val \\
\midrule
AE & 0.879 & \textbf{$<$0.001} & \textbf{0.935} & \textbf{$<$0.001} & 0.391 & 1.000 & 0.397 & 1.000 \\
BT-with-noise & 0.830 & 0.551 & 0.700 & 1.000 & 0.707 & 1.000 & 0.827 & 0.671 \\
\textbf{TwinPurify (TP)} & \textbf{0.991} & \textbf{$<$0.001} & 0.929& \textbf{$<$0.001} & \textbf{0.967} & \textbf{$<$0.001} & \textbf{0.905} & \textbf{$<$0.001} \\
VAE & 0.712 & 1.000 & 0.798 & 0.997 & 0.799 & 0.998 & 0.780 & 1.000 \\
\bottomrule
\end{tabular}
\end{table*}

To further investigate the functional specificity captured by each latent dimension, we generated pathway-level heatmaps for the top preranked genes per dimension (FWER~$<$ 0.01), filtering and aggregating pathways using GO semantic similarity (GOSemSim\cite{GOSEMSIM}) with a cutoff of 0.3 to reduce redundancy. Comparison across models (TP, AE, VAE; Fig.~\ref{fig:HEATMAP_SURVIVAL}a, Supplementary Fig.~S7, Supplementary Fig.~S8 ) reveals that TP consistently aggregates a larger number of pathways per dimension, both in total count and in the number of child terms per parent pathway. Although some overlap of pathways is observed across all models, likely reflecting inherent properties of the original gene expression data, TP more clearly delineates dimension-specific themes. Major dimensions in TP are enriched for immune response, cartilage development, DNA replication, and cell cycle-related pathways, indicating substantial functional independence among latent axes. While complete segregation of pathways per dimension is not fully achieved, this is expected given the high interdependence of biological processes and frequent gene overlap across pathways. These results demonstrate that TP embeddings effectively capture interpretable, dimension-specific transcriptional programs while respecting the inherent structure of biological networks.

\subsubsection{Survival prediction using latent gene sets}

\begin{figure*}[htbp]
  \centering
  \includegraphics[width=\linewidth]{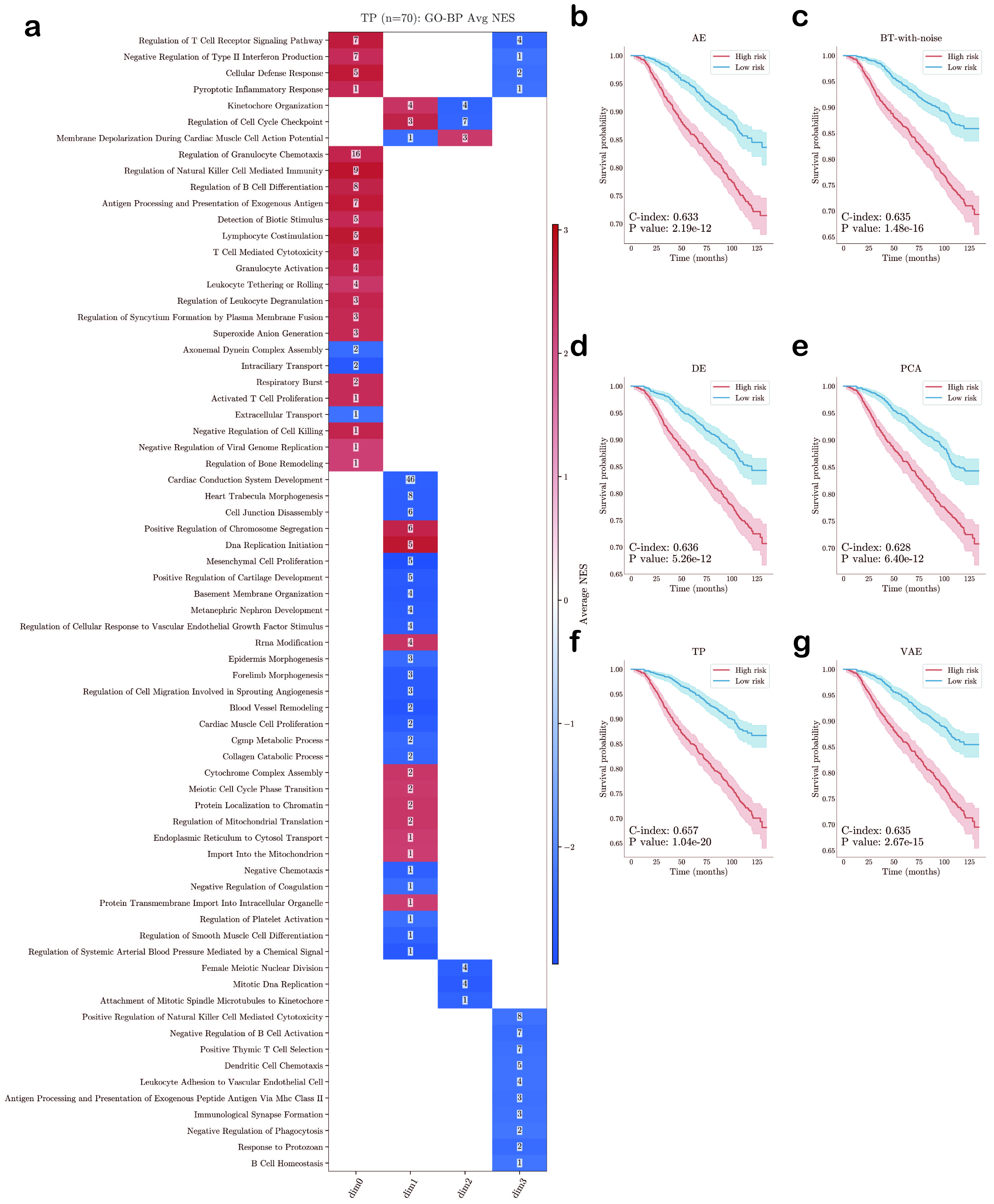}
  \caption{%
    \textbf{Functional relevance and survival association of embedding dimensions.}
    \textbf{a} Significantly enriched pathways (FWER $<$  0.01) identified from TwinPurify embeddings after summarization. Each cell shows the number of child terms within a summarized parent term, and color intensity indicates the normalized enrichment score (NES).
    \textbf{b-g} Kaplan-Meier survival analyses based on the top and bottom ranked genes (20 each per dimension) derived from preranked gene lists across six models, with 160 genes used per model.
  }
  \label{fig:HEATMAP_SURVIVAL}
\end{figure*}

Finally, we validated the clinical utility of the learned representations by quantifying the prognostic value of embedding-derived gene signatures using Cox proportional hazards modeling and Kaplan-Meier analysis (Fig.~\ref{fig:HEATMAP_SURVIVAL}b-g). TP consistently outperformed all other models, exhibiting higher C-index values and significantly lower p-values (by 5-8 orders of magnitude) compared to AE, VAE, and DE. The BT-with-Gaussian-noise model, which uses Gaussian noise as a perturbation instead of adjacent-normal tissue, achieved intermediate performance: better than AE/VAE but inferior to TP. These results indicate that incorporating adjacent-normal as a directed perturbation effectively purifies tumor-intrinsic signals, producing more clearly defined high- and low-risk groups and enhancing the predictive power of survival models.

Overall, TwinPurify embeddings reveal biologically coherent and largely independent transcriptional programs, capturing both tumor-intrinsic and immune-related signals. Gene sets derived from these dimensions also show superior prognostic value for patient survival compared with baseline models, highlighting the functional and clinical relevance of the purified tumor representations.

\section{Discussion and conclusion}

Extracting tumor-specific signals from bulk transcriptomic and microarray data remains challenging. This difficulty arises from contamination by adjacent-normal tissue and the inherent complexity of the tumor microenvironment. To address this, we developed TwinPurify (TP), a self-supervised framework inspired by the Barlow Twins architecture, designed to enhance tumor-intrinsic and immune signals while minimizing non-tumor contributions. TP learns representations that capture biologically meaningful and largely independent axes of variation, reflecting intrinsic cancer cell programs disentangled from background noise. Our analyses demonstrate that TP exhibits strong robustness to normal tissue contamination, extracts coherent gene- and pathway-level signals, and produces gene signatures that more effectively stratify patient outcomes compared to conventional approaches.

The conceptual strength of TP lies in its embedding design. The Barlow Twins-based architecture naturally aligns with the goal of disentangling orthogonal biological programs. Traditional deconvolution or statistical approaches process bulk gene expression as a single high-dimensional entity, resulting in pathway signals that are entangled into broad clusters rather than resolved into distinct biological processes. Even embedding-based models such as autoencoders (AE) and variational autoencoders (VAE) fall short of this objective, as they lack the decorrelation constraint that explicitly enforces independence among latent dimensions, which is a key requirement for revealing distinct transcriptional axes.

While the original Barlow Twins framework was developed for computer vision, where distortions such as cropping, color shifting, or scaling are used to create augmented image views, these perturbations are not interpretable or biologically meaningful for transcriptomic data. We demonstrate that generic Gaussian noise is insufficient for purifying tumor signals. In gene expression, random Gaussian noise does not capture biologically relevant variation; in contrast, using adjacent-normal profiles provides a natural “negative” supervision, explicitly teaching the model to recognize and suppress the specific transcriptional signature of healthy tissue. To adapt the framework to a biological context, TP replaces generic noise with adjacent-normal profiles as structured perturbations. This biologically grounded design leverages the contrast between tumor and adjacent-normal tissues as a natural form of supervision, enabling TP to purify cancer-intrinsic and microenvironmental signals rather than merely improving model robustness. 

Crucially, our results distinguish between the “background” signals of adjacent healthy tissue and the “active” signals of the tumor immune microenvironment. While TwinPurify is designed to subtract the expression profile of healthy adjacent tissue, it preserves and effectively enhances signals from tumor-infiltrating immune cells. This is likely because immune infiltration is a hallmark of the tumor ecosystem that is distinct from the quiescent background of adjacent normal tissue. By using adjacent-normal profiles as the negative reference, TP filters out the stromal/normal background while retaining the high-variance signals associated with immune responses.

Existing deconvolution methods such as MuSiC and TAPE depend on integrating single-cell and bulk RNA-seq datasets collected from separate samples, where protocol disparities are common. These methods implicitly assume that reference signatures are strictly transferable to target cohorts, an assumption that frequently fails in clinical reality due to biological and technical mismatches. Consequently, they often achieve strong performance on simulated mixtures but generalize poorly to real patient samples. In contrast, TP circumvents this issue by directly learning tumor-intrinsic and microenvironmental signals from adjacent bulk measurements, reducing dependency on external references and minimizing bias. This design enables TP to provide a robust and scalable alternative to transcriptomic deconvolution across cohorts and platforms, without necessitating explicit cell-type inference. This represents a fundamentally different approach that does not rely on explicit cell-type deconvolution. 

The generality of TP also suggests broad applicability beyond breast cancer. In principle, any identifiable contaminating cell-type (immune, stromal, or epithelial) could serve as a distortion source. By providing expression profiles of these confounding cell populations, one can generate cleaner representations of the cell type of interest and effectively “subtracting” biological noise without explicit supervision. Additionally, integrating TP with emerging data types such as single-cell RNA-seq or spatial transcriptomics may further enhance interpretability and validation, particularly where spatial coordinates or cell-type annotations are available to serve as partial ground truth.

Nonetheless, this work has limitations. Foremost, the lack of absolute ground truth for tumor purity hinders definitive benchmarking. Our dilution framework thus serves as a proxy evaluation strategy, enabling relative performance comparisons but not absolute quantification. Moreover, the observed decorrelation of embedding dimensions is advantageous for disentanglement, but it does not guarantee biological independence, as biological pathways are often interdependent and co-regulated.

In summary, we present a novel, resilient method for purifying bulk tumor gene expression that does not rely on labelled cell types or reference profiles. By leveraging adjacent-normal samples as structured perturbations, TP learns robust, interpretable embeddings that retain critical biological signals and support more accurate downstream inference. These findings offer a new perspective on unsupervised representation learning for bulk transcriptomic data and suggest future applications in precision oncology, cancer biology, and integrative omics. 

\begin{center}
\fbox{%
\begin{minipage}{0.95\linewidth}
\textbf{Key Points}

\begin{itemize}
  \item TwinPurify introduces \emph{computational purification}, a framework that mathematically disentangles and removes contaminating adjacent-normal tissue signals from bulk tumor expression profiles without requiring external reference samples.

  \item TwinPurify is the first application of the Barlow Twins methodology to disentangle distinct biological programs from cancer gene expression data.

  \item TwinPurify provides a user-friendly Python software package that enables researchers to easily implement the framework and generate more accurate, biologically coherent Gene Set Enrichment Analysis (GSEA) results.
  
  \item Extensive benchmarking across large-scale cohorts demonstrates that TwinPurify yields representations that are robust to purity dilution and consistently outperform baseline methods.
  
  \item The purified embeddings enhance downstream clinical utility by improving molecular subtyping, tumor grading, and survival prediction, while amplifying key tumor microenvironment signals, including immune signatures.
\end{itemize}

\end{minipage}}
\end{center}

\section{Competing interests}
No competing interest is declared.

\section{Author contributions statement}
Zhiwei Zheng (Conceptualization, Methodology, Software, Validation, Formal analysis, Investigation, Data curation, Visualization, Writing—original draft) and Kevin Bryson (Conceptualization, Methodology, Resources, Writing—review \& editing, Supervision, Project administration).

\section{Acknowledgments}
The authors thank the anonymous reviewers for their valuable suggestions. 

\section{Software and data availability}
The source code for TwinPurify, including data preprocessing scripts, model training pipelines, and downstream analysis notebooks, is available for peer review at \url{https://anonymous.4open.science/r/TwinPurify-6DF7}; upon acceptance, the complete codebase will be permanently archived on GitHub. The RNA-seq datasets analyzed in this study are publicly accessible: the SCAN-B dataset~\cite{scanb_dataset} is available via Mendeley Data (\url{https://doi.org/10.17632/yzxtxn4nmd.4}); the TCGA-BRCA dataset~\cite{ucsc_xena_tcga_gtex} was obtained from the UCSC Xena Browser (\url{https://xenabrowser.net/datapages/?cohort=TCGA%20TARGET%20GTEx&
removeHub=https%3A%2F%2Fxena.treehouse.gi.ucsc.edu%3A443.}); and the METABRIC dataset~\cite{ega_breast_cancer} can be accessed through the European Genome-phenome Archive (EGA) under accession number \textbf{EGAS00000000083} (\url{https://ega-archive.org/studies/EGAS00000000083}).


\bibliographystyle{unsrtnat}   
\bibliography{Reference}        

\end{document}